\documentclass[preprint,12pt,authoryear]{elsarticle}

\usepackage{amssymb}
\usepackage{amsthm}

\usepackage{enumitem}
\usepackage{amsmath}
\usepackage{makecell}
\usepackage{multirow}
\usepackage{adjustbox}
\usepackage{booktabs}
\usepackage{graphicx}
\usepackage{float}
\usepackage{xcolor}
\usepackage{mathtools}
\usepackage{etoolbox}
\usepackage{setspace}
\usepackage{units}
\usepackage{subcaption}
\usepackage{comment}
\usepackage[ruled,vlined]{algorithm2e}
\usepackage[hidelinks]{hyperref}

\newtheorem*{assumption*}{\assumptionnumber}
\providecommand{\assumptionnumber}{}
\makeatletter

\definecolor{red(ncs)}{rgb}{0.0, 0.0, 0.0}
\newcommand{\rd}[1]{\textcolor{red(ncs)}{#1}}

\DeclarePairedDelimiter\floor{\lfloor}{\rfloor}

\journal{}

\begin{document}

\begin{frontmatter}

%% Title, authors and addresses

%% use the tnoteref command within \title for footnotes;
%% use the tnotetext command for theassociated footnote;
%% use the fnref command within \author or \affiliation for footnotes;
%% use the fntext command for theassociated footnote;
%% use the corref command within \author for corresponding author footnotes;
%% use the cortext command for theassociated footnote;
%% use the ead command for the email address,
%% and the form \ead[url] for the home page:
%% \title{Title\tnoteref{label1}}
%% \tnotetext[label1]{}
%% \author{Name\corref{cor1}\fnref{label2}}
%% \ead{email address}
%% \ead[url]{home page}
%% \fntext[label2]{}
%% \cortext[cor1]{}
%% \affiliation{organization={},
%%            addressline={}, 
%%            city={},
%%            postcode={}, 
%%            state={},
%%            country={}}
%% \fntext[label3]{}

\title{Self-organized \rd{free-flight} arrival for urban air mobility}

%% use optional labels to link authors explicitly to addresses:
%% \author[label1,label2]{}
%% \affiliation[label1]{organization={},
%%             addressline={},
%%             city={},
%%             postcode={},
%%             state={},
%%             country={}}
%%
%% \affiliation[label2]{organization={},
%%             addressline={},
%%             city={},
%%             postcode={},
%%             state={},
%%             country={}}

\author[a]{Martin Waltz\corref{CorrespondingAuthor}}
\ead{martin.waltz@tu-dresden.de}

\author[a,b]{Ostap Okhrin}
\author[c]{Michael Schultz}

\affiliation[a]{organization={Technische Universität Dresden, Chair of Econometrics and Statistics, esp. in the Transport Sector}, city={Dresden}, postcode={01062, Wuerzburger Str. 35}, country={Germany}}

\affiliation[b]{organization={Center for Scalable Data Analytics and Artificial Intelligence (ScaDS.AI)}, city={Dresden/Leipzig}, country={Germany}}

\affiliation[c]{organization={Institute of Flight Systems, Universität der Bundeswehr München}, city={Neubiberg}, postcode={85577}, country={Germany}}

\cortext[CorrespondingAuthor]{Corresponding author}
\begin{abstract}
%% Text of abstract
Urban air mobility is an innovative mode of transportation in which electric vertical takeoff and landing (eVTOL) vehicles operate between nodes called vertiports. We outline a self-organized vertiport arrival system based on deep reinforcement learning. The airspace around the vertiport is assumed to be circular, and the vehicles can freely operate inside. Each aircraft is considered an individual agent and follows a shared policy, resulting in decentralized actions that are based on local information. We investigate the development of the reinforcement learning policy during training and illustrate how the algorithm moves from suboptimal local holding patterns to a safe and efficient final policy. The latter is validated in simulation-based scenarios\rd{, including robustness analyses against sensor noise and a changing distribution of inbound traffic.~Lastly, we deploy the final policy} on small-scale unmanned aerial vehicles to showcase its real-world usability.
\end{abstract}

%%Graphical abstract
%\begin{graphicalabstract}
%\includegraphics{grabs}
%\end{graphicalabstract}

%%Research highlights
%\begin{highlights}
%\item Research highlight 1
%%\end{highlights}

\begin{keyword}
deep reinforcement learning \sep urban air mobility \sep eVTOL
%% keywords here, in the form: keyword \sep keyword

%% PACS codes here, in the form: \PACS code \sep code

%% MSC codes here, in the form: \MSC code \sep code
%% or \MSC[2008] code \sep code (2000 is the default)

\end{keyword}

\end{frontmatter}

%% \linenumbers

%% main text

\sloppy
\section{Introduction}
\label{sec:introduction}
Urban air mobility (UAM) constitutes a solution to alleviate traffic congestion in urban centers by offering on-demand air transportation services \citep{mueller2017enabling}. The concept leverages electric vertical takeoff and landing (eVTOL) aircraft for sustainable and efficient passenger transport and additional tasks like emergency medical evacuations and package delivery \citep{thipphavong2018urban}. The surge of interest in UAM is evident in the development efforts by industry leaders such as Airbus, Boeing, and Volocopter, who are actively engaged in the design and testing of eVTOL vehicles \citep{polaczyk2019review}.

In contrast to terrestrial road traffic, UAM operates on a node-based transportation system \citep{thipphavong2018urban}. These nodes, known as vertiports, serve as centralized hubs equipped for custom drop-offs and pick-ups, maintenance, and the charging of eVTOL vehicles \citep{rajendran2020air}. UAM vehicles are anticipated to navigate between vertiports without hindrance, eliminating the necessity for fixed routes or rigid flight plans. However, insights from conventional airspace surveillance suggest that consolidating traffic reduces control workload and enhances system capacity. On this basis, \cite{bertram2020efficient} and \cite{kleinbekman2018evtol} argue that some form of traffic flow structure, such as the definition of terminal arrival gates, will likely be imposed. Moreover, \cite{kleinbekman2018evtol} state that the terminal arrival phase most likely constitutes the most safety-critical part of UAM operations, given the prospect of high-density terminal traffic coupled with limited landing capacities. While it is anticipated that some kind of terminal arrival air traffic control will be present, the vehicles are expected to autonomously respond to congestion by appropriately avoiding collisions \citep{bertram2020efficient}.

Therefore, the control approach for vehicles operating in the terminal airspace near a vertiport is a crucial practical concern, necessitating the construction of a carefully designed safety assurance and conflict resolution system. According to \cite{yang2020scalable}, such systems can be classified based on three criteria:
\begin{enumerate}
    \item \emph{Centralized/decentralized:} In a centralized approach, a supervising controller processes all available information and issues control commands to each aircraft. Conversely, in decentralized settings, each aircraft independently selects actions based on its local information.
    \item \emph{Planning/reacting:} Planning approaches define paths or trajectories before execution while reacting methods make online decisions based on the current positions of other aircraft or static obstacles.
    \item \emph{Cooperative/non-cooperative:} Cooperative scenarios entail explicit information exchange between aircraft, enabling communication. On the other hand, non-cooperative approaches restrict vehicles from sending or receiving messages from other aircraft.
\end{enumerate}

In contrast to decentralized methods, centralized approaches typically plan the entire trajectory of vehicles and can achieve globally optimal solutions. However, they quickly become infeasible for large systems due to prolonged computation times. Additionally, decentralized, reactive systems demonstrate greater resilience to single-point failures, increasing their attractiveness from a practical point of view \citep{pallottino2006probabilistic}. Moreover, cooperative communication can simplify the problem and enhance operational safety, as vehicles can actively share their intentions and desired behavior. However, in reality, establishing reliable communication channels with high robustness against hardware failures is difficult \citep{chen2017decentralized}, what forces the underlying control algorithm to function even in the event of a loss of communication signal.

%We emphasize that \cite{yang2020scalable} refer to the third category as \emph{cooperative/non-cooperative}. However, motivated from the multi-agent reinforcement learning literature, where agents can cooperate to achieve a common goal without communication \citep{foerster2016learning, lowe2017multi}, we prefer the classification \emph{communicative/non-communicative}.

Consolidating these ideas, this paper introduces a decentralized, reactive, and non-cooperative vehicle control method within the terminal airspace surrounding a vertiport, using deep reinforcement learning (DRL, \citealt{arulkumaran2017deep}). DRL combines the expressive power of deep learning \citep{lecun2015deep} with the trial-and-error learning paradigm of reinforcement learning (RL, \citealt{sutton2018reinforcement}). RL represents a branch of artificial intelligence where an agent interacts with an environment to maximize a reward. Given its broad applicability to sequential decision tasks, this method has demonstrated remarkable results across diverse domains, including autonomous driving \citep{feng2023dense}, molecular optimization \citep{zhou2019optimization}, portfolio management \citep{hu2019deep}, and even algorithmic breakthroughs such as the discovery of efficient matrix multiplication methods \citep{fawzi2022discovering} and improved sorting algorithms \citep{mankowitz2023faster}.

Similar to \cite{bertram2020efficient}, we define a circular airspace design around a vertical takeoff and landing (VTOL) zone, with multiple aircraft freely operating in the airspace. Each aircraft is considered a separate RL agent, constituting a multi-agent reinforcement learning (MARL, \citealt{zhang2021multi}) scenario. There are different approaches to designing the specifics of a MARL problem, depending on the information level, objective definition, and communication capabilities of the agents \citep{wang2022review}. In our pursuit of a decentralized, reactive, and non-cooperative approach to separation assurance, we specify \emph{one policy} that is \emph{shared across agents}, enabling training in a \emph{single-agent} fashion. The shared policy uses a recurrent neural network architecture recently proposed in the maritime traffic domain \citep{waltz2023spatial}. This approach allows to handle an arbitrary number of aircraft in the airspace while ensuring the eVTOL vehicles take decentralized actions based on their local observations.

In summary, the contributions of this work are as follows:
\begin{itemize}
    \item We formulate a circular airspace design for the terminal arrival of eVTOL vehicles under the free-flight framework \citep{hoekstra2002designing}, and outline a DRL-based approach to autonomously organize the traffic. The final policy results in safe and efficient operations.
    \item In the process, we define the observation, action, and reward spaces for optimizing the shared policy. Moreover, we use curriculum learning \citep{narvekar2020curriculum} to gradually increase the complexity of training scenarios, strongly improving the effectiveness of the final policy.
    \item We conduct testing through scenario-based validation, thorough simulation \rd{studies, and different robustness analyses}. Moreover, we investigate and visualize how the policy changes during training, delivering further insights into the learning process.
    \item \rd{Crucially}, we demonstrate the effectiveness of our policy, entirely trained in simulation, through real-world experiments conducted on small-scale unmanned aerial vehicles called \emph{Crazyflies} \citep{giernacki2017crazyflie}. The demonstrated success of the Sim-2-Real transfer highlights the practical applicability of our results.
\end{itemize}
The paper is structured as follows: Section \ref{sec:related_work} reviews related work regarding safety assurance systems and terminal arrival organization in UAM operations. Section \ref{sec:problem} defines the airspace design and describes the RL-based modeling approach. Subsequently, Section \ref{sec:solution} details the definition of the observation, action, and reward spaces, the algorithmic background, and the training environment. Afterward, Section \ref{sec:results} displays the results, \rd{while Section \ref{sec:robustness} contains the robustness analysis,} including the real-world experiments. Section \ref{sec:conclusion} concludes this paper. The source code to this work is publicly available at the GitHub repository \cite{TUDRL}.

\section{Related work}\label{sec:related_work}
\cite{straubinger2020overview}, \cite{garrow2021urban}, and \cite{rajendran2020air} provide reviews about recent research and current developments in urban air mobility operations. In particular, recent studies often focus on demand forecasting for UAM services \citep{wu2021integrated, mayakonda2020top}, airspace design and integration concepts \citep{UberWhitePaper, thipphavong2018urban, bauranov2021designing}, and vehicle design \citep{silva2018vtol, brown2020vehicle}.

More closely connected to our research, there have been recent proposals for separation assurance and collision avoidance systems in UAM environments. \cite{yang2020scalable} propose a multi-agent computational guidance algorithm for on-demand air transportation. The work is based on a Monte Carlo tree search (MCTS) and incorporates an asynchronous message-passing scheme, enabling agents to include information about other agents' actions in their action selection procedure. However, \cite{yang2020scalable} rely on a simple discrete action space, whose expansion is difficult due to the exponential growth of the search tree with the number of actions. Similar contributions based on MCTS include \cite{wu2022safety} and \cite{yang2021autonomous}. Another sampling-based approach was introduced by \cite{wu2022risk}, which builds on the rapidly-exploring random tree \citep{lavalle1998rapidly} path planning algorithm. Moreover, the study introduces probabilistic risk bounds for collisions by assuming that aircraft positions follow a Gaussian distribution.

While DRL has gained increasing popularity in research on conventional air traffic control methodologies \citep{zhao2021physics,  wang2022review, brittain2022scalable}, its application for UAM separation assurance is strongly limited. A notable contribution is \cite{jang2020learning}, which builds on the Learning-to-Fly framework of \cite{rodionova2020learning} for collision avoidance. These works consider a communicative scenario with an explicit exchange of planned trajectories between UAM aircraft. Conflict resolution then takes place based on a combination of supervised learning \citep{rodionova2020learning} or reinforcement learning \citep{jang2020learning}, respectively, and model predictive control. Being remotely connected, \cite{huang2023strategic} outline an approach for strategic conflict management based on MARL. The authors construct spatial-temporal UAM flight trajectories for a set of vertiports and apply MARL to choose between ground delay, speed adjustment, or cancellation for conflicting flights. In addition, \cite{park2023multi} outline a UAM transportation service management system for vehicles operating between multiple vertiports. Methodologically, the authors rely on MARL in the form of the CommNet approach of \cite{sukhbaatar2016learning}, explicitly allowing for communication between agents.

Furthermore, some selected works focus on eVTOL aircraft arrival optimization. In particular, \cite{kleinbekman2018evtol} use a mixed-integer linear program formulation to compute the optimal required time of arrival (RTA) for eVTOL aircraft. In addition, the authors propose a concept of operations for vertiport terminal area airspace design. The design includes a vertiport with two arrival and two departure metering fixes, allowing to separate descending and climbing traffic. \cite{kleinbekman2020rolling} extend the work of \cite{kleinbekman2018evtol} by considering double-landing-pad vertiports and a rolling-horizon, mixed-integer linear program formulation for the RTA optimization. The computed RTAs can serve as a basis for the proposal of \cite{pradeep2019energy}, where a multiphase optimal control framework is presented that performs energy-efficient eVTOL aircraft arrival for given RTAs. Finally, \cite{song2021development} introduce and compare different strategies for scheduling arriving aircraft in UAM operations, which are optimized using a genetic algorithm \citep{whitley1994genetic}.

Closest to our work is \cite{bertram2020efficient}, which uses the algorithm outlined in \cite{bertram2020distributed} to perform separation assurance and collision avoidance for a UAM terminal arrival sequencing problem. In particular, the authors consider a terminal arrival airspace design consisting of rings with possibly different altitudes. Aircraft must sequentially pass from the outer to the inner rings until the vertiport landing zone is reached. The number of rings and traveling directions are explicitly specified, imposing a strict structure on the airspace.

\section{Problem statement}\label{sec:problem}
\subsection{Airspace design}\label{sec:airspace_design}
\begin{figure}[ht]
    \centering
    \includegraphics[width=0.7\textwidth]{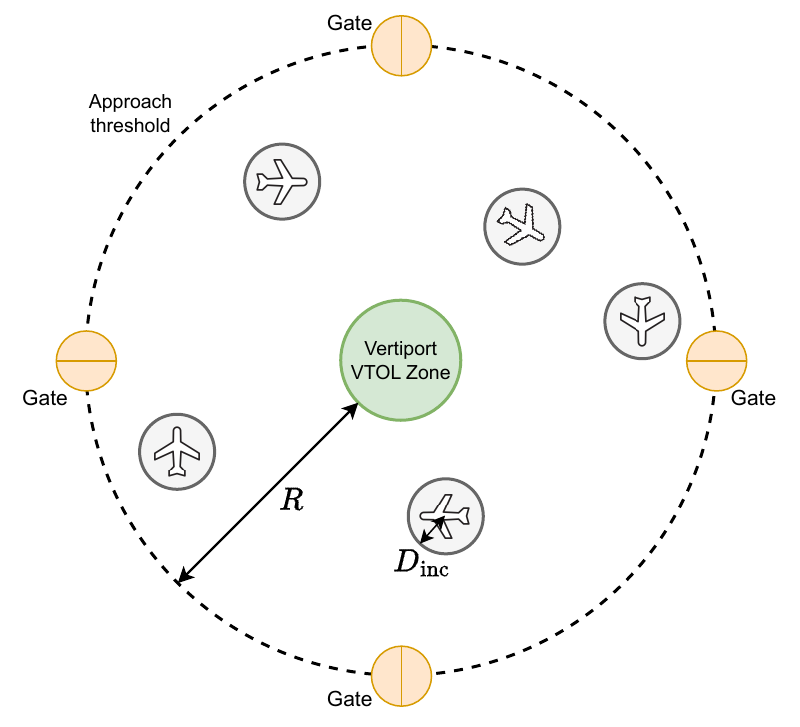}
    \caption{Schematic illustration of the airspace design; inspired by \cite{bertram2020efficient}.}
    \label{fig:airspace}
\end{figure}

We consider a circle-shaped terminal airspace around a vertiport, which is visualized in Figure \ref{fig:airspace}.~The vertiport VTOL zone is assumed to have a radius of $\unit[200]{m}$, followed by an airspace with an outer radius of $R = \unit[800]{m}$. \rd{As stated in \cite{easa}, safety regions with obstacle clearance will surround vertiports. Hence, we consider the airspace to be free of static obstacles, which would impact the vehicles' flight trajectories.} Following \cite{bertram2020efficient}, the incoming traffic is assumed to enter the airspace through one of the four gates, which are located in the north, east, south, and west directions from the vertiport. Afterward, the aircraft can freely operate inside the airspace, resembling free-flight concepts of conventional aircraft traffic \citep{hoekstra2002designing, ribeiro2022using, kleinbekman2018evtol, groot2024analysis}. The vehicles are \rd{assumed to be} unable to hover, possibly travel at different speeds, and can only adjust their respective heading. These assumptions, though restrictive, significantly elevate the complexity of the problem. \rd{In particular, while eVTOL vehicles typically have the ability to hover, the energy consumption is significantly higher than in the flight mode \citep{kasliwal2019role}, where horizontal speed additionally results in dynamic lift. From a conventional air traffic management perspective, our approach can thus be considered as a combination of holding patterns \citep{icao} and the point merge concept (semi-circle approach area; \citealt{eurocontrol}). In particular, our airspace design effectively adjusts the point merge concept to a circular area.}

\rd{Contrary} to \cite{bertram2020efficient}, we do not impose further constraints like pre-defined traffic rings inside the airspace or particular traveling directions of individual vehicles. Hence, the clock-wise direction of the five shown aircraft shown in Figure \ref{fig:airspace} serves purely as an illustration. In addition, \rd{similar to conventional air traffic, we assume that flight levels will separate arrival and departure flows. In particular, all arrival vehicles operate on the same altitude, avoiding interactions with departing aircraft at lower altitude.}

The objective of each aircraft is twofold. First, after entering the area, it should stay in the airspace without colliding with other vehicles by adjusting its heading. While altitude changes can be deployed in practice as emergency measures for collision avoidance, we do not consider them as regular control options within this context. Furthermore, we differentiate between accidents, which we consider as events where the distance between two vehicles is smaller than $D_{\rm acc} = \unit[10]{m}$, and incidents, which are similarly defined with a distance of $D_{\rm inc} = \unit[100]{m}$. Secondly, the aircraft should approach and enter the VTOL zone safely if the vertiport has available capacity, and it is the respective aircraft's turn. Crucially, only one vehicle is permitted to enter the VTOL zone at a time for a landing maneuver. During this period, the vertiport is temporarily blocked and inaccessible to other vehicles. Throughout the paper, we assume that the required time for such landing maneuvers is $T_{\rm land} = \unit[60]{s}$, which approximately aligns with the average time phasing of conventional aircraft inbound traffic.

A simple rule determines the selection process for the entering aircraft in our validation experiments: Priority has the aircraft with the smallest Euclidean distance to the VTOL zone. This rule can be implemented in our fully decentralized setting since each aircraft senses the surrounding vehicles' positions and can infer their respective vertiport distances. Suppose additional information, such as the time spent in the terminal airspace or the battery charging level of the vehicles, is accessible. In that case, the rule can be substituted with a first-in-first-out scheme, as outlined in \cite{bertram2020efficient}, or a lowest-battery approach. The flexibility of the selection process allows for adaptation based on the available data and specific operational requirements.

For illustrative purposes, we have chosen to simulate the behavior of the DJI Mavic Pro drone, whose dynamics are implemented in the Bluesky simulator developed by \cite{hoekstra2016bluesky}. We thereby set the simulation step size to $\Delta t = \unit[1]{s}$.

\subsection{Modeling approach}\label{subsec:modelling_approach}
At the time $t$, the terminal airspace accommodates $N_t$ aircraft, with this count dynamically changing over time as aircraft enter and depart. Each aircraft is treated as a distinct agent when solving this problem using multi-agent reinforcement learning (MARL). Adhering to the classification in \cite{wang2022distributed}, there are three fundamental paradigms to define the learning process: Centralized learning, independent learning, and centralized training decentralized execution (CTDE). In centralized learning, a single controller processes the observations of all agents and yields a joint vector of actions \citep{sukhbaatar2016learning}. Such an approach is undesirable since we aim for a robust decentralized solution to separation assurance. In independent learning, each agent optimizes for a separate policy and is trained in a single-agent fashion \citep{tan1993multi,tampuu2017multiagent}. However, this approach proves unsuitable for our scenario, as it typically assumes a constant number of agents. Lastly, CTDE combines the advantages of the first two categories since the execution remains distributed, but the learning phase is in a centralized setting \citep{foerster2016learning, lowe2017multi}. 

Our solution falls into the CTDE category since we use parameter sharing \citep{tan1993multi, gupta2017cooperative} across the agents. More precisely, we assume that each of the $N_t$ aircraft follows the \emph{same shared policy} but takes \emph{decentralized actions} based on its local observation. The shared policy uses a recurrent neural network to cope with a varying number of aircraft \citep{waltz20232}. We emphasize that this single-policy multiple-agent strategy constitutes an instance of curriculum learning \citep{narvekar2020curriculum}. Curriculum learning is a concept inspired by educational theory, describing sequential learning from tasks with increasing complexity. In our case, the curriculum contains two components. First, each agent interacts with agents exhibiting identical behavior within the same environment, a manifestation of self-play. This concept, notably successful in games such as Backgammon \citep{tesauro1995temporal} and Go \citep{silver2016mastering}, facilitates robust learning and adaptability. Second, we systematically elevate the complexity of scenarios throughout training by incrementally increasing the number of aircraft in the airspace. The detailed schedule is described in Section \ref{subsec:scenario_generation}.

\section{Solution method}\label{sec:solution}
\subsection{Reinforcement learning}
The basic model for formalizing the agent-environment interaction in RL is the Markov Decision Process (MDP, \citealt{puterman2014markov}), which is defined as the tuple $(\mathcal{S}, \mathcal{A}, \mathcal{P}, \mathcal{R}, \gamma)$. In this context, $\mathcal{S}$ signifies the state space, $\mathcal{A}$ denotes the action space, $\mathcal{P}: \mathcal{S} \times \mathcal{A} \times \mathcal{S} \rightarrow [0,1]$ is the state transition probability function, $\mathcal{R}: \mathcal{S} \times \mathcal{A} \rightarrow \mathbb{R}$ represents a bounded reward function, and $\gamma \in [0,1)$ serves as a discount factor, modulating the trade-off between immediate and future rewards. At time step $t$, the agent receives the state $s_t \in \mathcal{S}$, selects an action $a_t \in \mathcal{A}$, and transitions according to the dynamics $\mathcal{P}$ to a new state $s_{t+1}$. It thereby collects a reward $r_t$, generated by $\mathcal{R}(s_t, a_t)$. The objective of the agent is to optimize for a policy $\pi : \mathcal{S} \times \mathcal{A} \rightarrow [0,1]$, a mapping from states to actions, that maximizes a performance measure such as the expected discounted sum of future rewards $\operatorname{E}_{\pi}\left[\sum_{t=0}^{\infty}\gamma^t r_t\right]$.

Reinforcement learning algorithms are commonly classified into two main categories: Value-based and policy gradient approaches \citep{sutton1999policy}. In value-based approaches, the focus is often on estimating the action-value, commonly referred to as the $Q$-value, for a specific state-action pair $(s \in \mathcal{S}, a \in \mathcal{A})$ under a given policy $\pi$. This $Q$-value is defined as $Q^{\pi}(s,a) = \operatorname{E}_{\pi}\left[\sum_{t=0}^{\infty}\gamma^t r_t \vert s_0 = s, a_0 = a\right]$, where it represents the expected cumulative reward, discounted by $\gamma$, starting from the initial state-action pair $(s, a)$ and following the policy $\pi$. Importantly, under the fulfillment of regularity conditions, an optimal policy $\pi^*$ exists that is connected with optimal action-values $Q^*$ \citep{puterman2014markov}. Using recursive relationships originating from the dynamic programming literature allows to derive iterative schemes to approximate $Q^*$ and hence optimize for $\pi^*$ \citep{bellman1954theory}. On the other hand, policy gradient approaches specify a parametrized and differentiable policy, represented, for example, by a neural network. The policy parameters are directly optimized using the gradient of the performance measure \citep{sutton1999policy, silver2014deterministic}. An important class of policy gradient methods are actor-critic algorithms, which use an action-value estimate during the policy gradient update \citep{fujimoto2018addressing, haarnoja2018soft}. In this context, the policy is termed the \emph{actor}, while the action-value estimator is the \emph{critic} \citep{sutton2018reinforcement}.

However, the MDP model assumes to observe a physical system's true state, which is often unrealistic since real-world disturbances introduce noise to the sensed information or cause time delays in the measurements \citep{molchanov2019sim}. The formal extension of the MDP that handles such scenarios is the Partially Observable Markov Decision Process (POMDP, \citealt{littman2009tutorial}), which is defined as the tuple $(\mathcal{S}, \mathcal{A}, \mathcal{P}, \mathcal{R}, \gamma, \mathcal{O}, \mathcal{Z})$. It extends the MDP by introducing the observation space $\mathcal{O}$ and the observation function $\mathcal{Z}: \mathcal{S} \times \mathcal{A} \times \mathcal{O} \rightarrow [0,1]$. In contrast to an MDP, the agent-environment interaction in a POMDP is modified as follows: At each time step $t$, instead of observing the state $s_t$, the agent receives an observation $o_t \in \mathcal{O}$, which is generated by $\mathcal{Z}$. After selecting an action $a_t$, the system transitions to a state $s_{t+1}$, and the agent receives a new observation $o_{t+1}$. Hence, a POMDP can be understood as a Hidden Markov Model with actions.

\subsection{Observation and action space}\label{subsec:obs_act_space}
In this work, a local observation of an aircraft at time $t$, $o_t$, consists of two components:
\begin{equation}
    o_t = \left(\left(o_{\text{O},t}\right)^\top, \left(o_{\text{T},t}\right)^\top \right)^\top,
\end{equation}
where $o_{\text{O},t}$ contains information about the vehicle itself and $o_{\text{T},t}$ yields information about the target vehicles that surround the aircraft. More precisely, we set:
\begin{equation}
    o_{\text{O},t} = \left(\frac{\left[\alpha_{\text{O},t}^{\text{V}}\right]^\pi_{-\pi}}{\pi}, \frac{d_{\text{O},t}^{\text{V}}}{d_{\rm scale}}, \sigma_{\text{O},t} \right)^\top,
\end{equation}
where $\alpha_{\text{O},t}^{\text{V}}$ is the relative bearing of the vertiport midpoint from the perspective of the own vehicle, $d_{\text{O},t}^{\text{V}}$ denotes the Euclidean distance between the own vehicle and the vertiport midpoint, and $d_{\rm scale} = \unit[1000]{m}$ is a normalizing constant. The binary variable $\sigma_{\text{O},t} \in \{-1,1\}$ constitutes the interface to the selection of the next landing aircraft, which is determined based on the Euclidean distance in the validation experiments. It takes the value 1 if the vehicle should enter the vertiport and the value $-1$ otherwise. The angle transformation $[\cdot]_{-\pi}^{\pi}: \mathbb{R} \rightarrow [-\pi, \pi)$ is defined in \cite{benjamin2017autonomous} as follows:
\begin{equation}
\left[\theta\right]_{-\pi}^{\pi} = 
\begin{cases} 
      \theta - \floor*{\frac{\theta +\pi}{2\pi}} \cdot 2\pi & \quad \text{if } \theta \geq 0, \\
      \theta + \floor*{\frac{-\theta +\pi}{2\pi}}\cdot 2\pi
      & \quad \text{if } \theta < 0, \\
\end{cases}
\end{equation}
where the floor operator $\floor*{\theta}$ returns the smallest integer smaller $\theta$. Regarding the observation about surrounding vehicles, we define:
\begin{equation}\label{eq:o_TS}
    o_{\text{T}, t} = \left(\left(o_{\text{T},1,t}\right)^\top, \ldots, \left(o_{\text{T},N_t-1,t}\right)^\top \right)^\top,
\end{equation}
where $N_t-1$ is the number of other vehicles in the airspace at time step $t$. The component $o_{\text{T},i,t}$, for $i = 1, \ldots, N_t-1$, contains information about the surrounding vehicle $i$ at time $t$, and is defined as follows:
\begin{equation}
    o_{\text{T}, i, t} = \left(
    \frac{d_{\text{O},t}^{\text{i}}}{d_{\rm scale}},
    \frac{\left[\alpha_{\text{O},t}^{\text{i}}\right]^\pi_{-\pi}}{\pi}, 
    \frac{v_{i,t} - v_{\text{O},t}}{v_{\rm scale}},  
    \frac{\left[\psi_{i,t}-\psi_{\text{O},t} \right]^{\pi}_{-\pi}}{\pi},
    \frac{d_{t, \text{CPA}}^{i}}{\Tilde{d}_{\rm scale}},
    \frac{t_{t, \text{CPA}}^{i}}{t_{\rm scale}}
    \right)^\top,
\end{equation}
where $d_{\text{O},t}^{\text{i}}$ is the Euclidean distance between the own vehicle and target vehicle $i$, and $\alpha_{\text{O},t}^{\text{i}}$ is the relative bearing of target vehicle $i$ from the perspective of the own vehicle. The variables $v_{i,t}, v_{\text{O},t}, \psi_{i,t}, \psi_{\text{O},t}$ denote the speed and heading of the target vehicle $i$ and the own vehicle, respectively. To quantify the risk of collision, the observation includes $d_{t, \text{CPA}}^{i}$ and $t_{t, \text{CPA}}^{i}$, which denote the distance and time to the closest point of approach (CPA, \citealt{julian2019deep}) of the own vehicle with target vehicle $i$. Although these two metrics rely on a linear trajectory prediction of the involved vehicles, we empirically found their inclusion beneficial for the agents' performance. We set the normalizing constants to $v_{\rm scale} = \unit[6]{m/s}$, $\Tilde{d}_{\rm scale} = \unit[100]{m}$, and $t_{\rm scale} = \unit[60]{s}$, and sort (\ref{eq:o_TS}) according to descending distances to the own vehicle.

The action space $\mathcal{A}$ is one-dimensional and continuous. In particular, an action at time $t$, $a_t \in [-1,1]$, changes the heading of the respective vehicle:
\begin{equation}
    \psi_{\text{O},t+1} = [\psi_{\text{O},t} + a_t \cdot \Delta]_{-\pi}^{\pi},
\end{equation}
where $\Delta = 5^\circ$.

\subsection{Reward function}
The reward function defines the feedback for an agent's action and thus dictates the final learned behavior. We identified four components that are essential to achieve a reasonable traffic flow. The first component, $r_{\text{coll},t}$, is a collision penalty since operational safety has the utmost priority. We define:
\begin{equation}\label{eq:r_coll_exp}
r_{\text{coll},t} = 
    \begin{cases}
      -10 & \text{for} \quad D_{\text{min},t} \leq D_{\text{inc}},\\
      c_1 \cdot \exp{}\left[-(D_{\text{min},t}-D_{\text{inc}})^2 /c_{2}^2\right] & \rm else,
   \end{cases}
\end{equation}
where $D_{\text{min},t} = \min_{i=1,\ldots,N_t-1} d_{\text{O},t}^{\text{i}}$ is the distance to the closest other vehicle, $D_{\text{inc}} = \unit[100]{m}$ is the incident distance, and $c_1 = -5$ and $c_2 = \unit[160.5]{m}$ are constants. The latter have been chosen that the exponential term in (\ref{eq:r_coll_exp}) takes the value $-5$ if $D_{\text{min},t} = \unit[100]{m}$ and approximately value $-0.01$ if $D_{\text{min},t} = \unit[500]{m}$, ensuring the vehicles keep sufficient safety distance to each other in the airspace. 

The second reward component, $r_{\text{goal},t}$, directs a vehicle towards the vertiport if it is the vehicle's turn to enter the VTOL zone ($\sigma_{\text{O},t}=1$). In addition, we add a penalty if the vehicle enters the eVTOL zone, although it currently is not supposed to do so ($\sigma_{\text{O},t}=-1$). Formalizing these thoughts, we define:
\begin{equation}
r_{\text{goal},t} = 
    \begin{cases}
      (d_{\text{O},t-1}^{\text{V}}-d_{\text{O},t}^{\text{V}})/c_4 & \text{for} \quad \sigma_{\text{O},t}=1,\\
      -5 & \text{for} \quad (\sigma_{\text{O},t}=-1) \land (d_{\text{O},t}^{\text{V}} \leq c_5),\\
      0 & \rm else,
   \end{cases}
\end{equation}
where $c_4 = \unit[10]{m}$ and $c_5 = \unit[200]{m}$ are constants. The third reward component is denoted $r_{\text{space},t}$ and ensures the vehicles do not leave the considered airspace:
\begin{equation}
r_{\text{space},t} = 
    \begin{cases}
      -5 & \text{for} \quad d_{\text{O},t}^{\text{V}} \geq \unit[1000]{m},\\
      0 & \rm else.
   \end{cases}
\end{equation}
Lastly, the fourth component, $r_{\text{comf},t} = -(a_t)^4$, is a comfort reward to avoid overly frequent heading changes and encourage smooth control behavior. On this basis, the total reward at time $t$ is constructed as a linear combination of the four components:
\begin{equation}
    r_t = \omega_{\text{coll}} \cdot r_{\text{coll},t} + \omega_{\text{goal}} \cdot r_{\text{goal},t} + \omega_{\text{space}}\cdot r_{\text{space},t} + \omega_{\text{comf}} \cdot r_{\text{comf},t},
\end{equation}
where the weights $\omega_{\text{coll}} = \omega_{\text{goal}} = \frac{3}{7}$, $\omega_{\rm space} = \omega_{\rm comf} = \frac{2}{7}$ have been identified experimentally via a grid search.

\subsection{Algorithm}\label{subsec:algorithm}
\begin{figure}[htp]
    \centering
    \includegraphics[width=\textwidth]{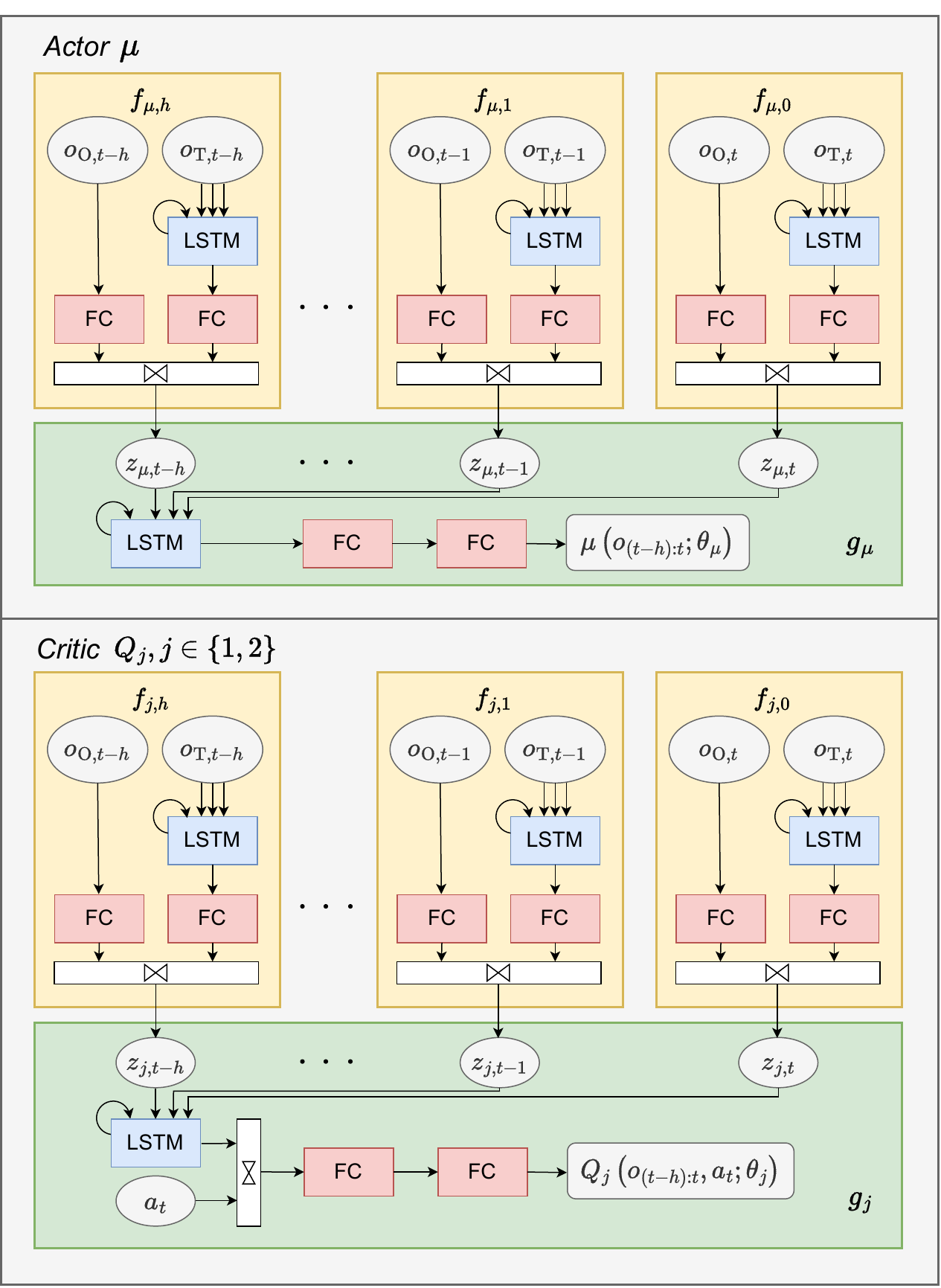}
    \caption{Neural network architecture adapted from \cite{waltz20232}. The symbol $\bowtie$ denotes concatenation.}
    \label{fig:LSTMRecTD3}
\end{figure}

In this study, we employ the LSTM-TD3 algorithm of \cite{meng2021memory}, which processes observations of multiple time steps through long short-term memory (LSTM, \citealt{hochreiter1997long}) layers. The algorithm extends the TD3 method of \cite{fujimoto2018addressing} to offer robustness against partial observabilities. Like the TD3, the LSTM-TD3 is an actor-critic algorithm using an actor network $\mu$ for the policy and two critics $Q_1$ and $Q_2$ for action-value estimation. Building on the LSTM-TD3 framework, \cite{waltz2023spatial} have proposed a spatial-temporal recurrent neural network architecture capable of handling information from a variable number of surrounding vehicles. Initially developed for the maritime domain, we adapt this network architecture to our specific context involving eVTOL vehicles. Given that \cite{waltz2023spatial} focus on discrete action spaces, we adopt the modification outlined in \cite{waltz20232}, which is designed for continuous action spaces. Figure \ref{fig:LSTMRecTD3} visualizes the architecture. Each fully connected (FC) layer in the figure uses 64 neurons, while the LSTM layers have 64 hidden units.

Formally, the actor $\mu$ is a neural network described as follows:
\vspace{-0.2cm}
\begin{align}
    z_{\mu, t-l} &= f_{\mu,l}\left(o_{\text{O},t-l}, o_{\text{T},t-l}; \theta_{f_{\mu,l}}\right) \quad \text{for} \quad l = 0, \ldots, h,\\
    \mu\left(o_{(t-h):t}; \theta_{\mu}\right) &= g_{\mu}\left(z_{\mu,t-h}, \ldots, z_{\mu,t-1}, z_{\mu,t}; \theta_{g_{\mu}}\right),\nonumber
\end{align}
where the functions $f_{\mu,l}$ with parameter sets $\theta_{f_{\mu,l}}$ for $l=0, \ldots,h$ represent the spatial recurrent components, which loop over the surrounding vehicles in the airspace. The function $g_{\mu}$, parametrized by $\theta_{g_{\mu}}$, represents the temporal recurrency since it aggregates information of the past $h$ time steps, in addition to the current information of time step $t$. Thus, the actor is a function of $h + 1$ observations, which is emphasized by the notation $o_{(t-h):t} = \cup_{l=0}^{h}o_{t-l}$. The overall parameter set of the actor is denoted $\theta_{\mu} = \left(\cup_{l=0}^{h} \theta_{f_{\mu,l}}\right) \cup \theta_{g_{\mu}}$.

Furthermore, the critic $Q_j$ with $j \in \{1,2\}$ can be formally expressed as follows:
\vspace{-0.4cm}
\begin{align}
    z_{j, t-l} &= f_{j,l}\left(o_{\text{O},t-l}, o_{\text{T},t-l}; \theta_{f_{j,l}}\right) \quad \text{for} \quad l = 0, \ldots, h,\\
    Q_j\left(o_{(t-h):t}, a_{t}; \theta_{j}\right) &= g_{j}\left(z_{j,t-h}, \ldots, z_{j,t-1}, z_{j,t}, a_{t}; \theta_{g_{j}}\right),\nonumber
\end{align}
Here, the spatial recurrent functions $f_{j,l}$ for $l = 0, \ldots, h$ are parameterized with sets $\theta_{f_{j,l}}$, and the function $g_j$ with parameter set $\theta_{g_j}$ represents the temporal recurrent component. The critics assess the action $a_{t}$ provided by the actor, and the complete parameter set of critic $Q_j$ is denoted as $\theta_{j} = \left(\cup_{l=0}^{h} \theta_{f_{j,l}}\right) \cup \theta_{g_j}$. Throughout the paper, we set $h=2$ for both the actor and the critics.

\subsection{Training scenario generation}\label{subsec:scenario_generation}
In the following, we denote a Bernoulli distribution with probability $p$ as $\mathcal{B}(p)$, a uniform distribution with the support range $[a,b]$ as $\mathcal{U}(a,b)$, and a discrete uniform distribution on the interval $[a,b]$ as $\mathcal{DU}(a,b)$. As mentioned in Section \ref{subsec:modelling_approach}, we pursue a curriculum learning strategy \citep{narvekar2020curriculum} to increase the number of vehicles in the airspace gradually. For the first $10^6$ training steps, we randomly sample $\mathcal{DU}(3,8)$ aircraft; for the subsequent $5 \cdot 10^5$ steps we randomly generate $\mathcal{DU}(8,15)$ vehicles; and for the remaining $5 \cdot 10^5$ steps we randomly sample the number of aircraft from $\mathcal{DU}(15,25)$.

To further enhance the diversity of encountered situations, we relax the gate-entry assumption during training. We initialize aircraft around the entire approach threshold illustrated in Figure \ref{fig:airspace} by sampling a corresponding entrance angle from $\mathcal{U}(0,360) \cdot 1^\circ$. Each aircraft is thereby generated with a heading pointing to the center of the VTOL zone, to which we add noise sampled from $(-1)^{\mathcal{B}(0.5)} \cdot \mathcal{U}(20, 45) \cdot 1^\circ$. The speed of an aircraft is sampled from $\mathcal{U}(10,16)\cdot \unit[1]{m/s}$. If an aircraft is more than $\unit[1200]{m}$ away from the VTOL zone midpoint during an episode, it is reinitialized.

\subsection{Experience replay}
Several state-of-the-art reinforcement learning algorithms, including the LSTM-TD3 introduced in Section \ref{subsec:algorithm}, rely on experience replay \citep{lin1992self}. Experience replay is a learning technique of randomly sampling past experiences from a replay buffer. The experiences are thereby stored in the form of transition tuples $(o,a,r,o')$, where $a \in \mathcal{A}$ is the selected action based on observation $o \in \mathcal{O}$, and $r$ and $o'$ denote the reward and the observation that followed. On this basis, an agent learns by periodically revisiting the experience it made. Although the approach allows for high data efficiency and training stability, the composition of the replay buffers is crucial \citep{hart2024enhanced}. If there are multiple tasks to learn or different situations to handle, the replay buffer should contain sufficiently diverse tuples to avoid over- or underrepresentations of particular scenarios \citep{chan2022zipfian}. A typical example of this issue is the common underrepresentation of experience associated with extreme braking in autonomous driving tasks \citep{hart2024towards}.

In our case, each aircraft has to perform two tasks: Safely staying in the airspace, corresponding to $\sigma_{\text{O},t} = -1$, and moving towards the VTOL zone, which translates to $\sigma_{\text{O},t} = 1$. Thus, the replay buffer should contain sufficient tuples associated with each task. To account for this, \emph{during training}, we randomly select one of the $N_t$ agents in the airspace as the main agent, whose experience is used for optimizing the shared policy. For the initial 200 steps of each training episode, all aircraft are instructed \emph{not} to enter the VTOL zone ($\sigma_{\text{O},t} = -1$). Subsequently, only the main agent receives an entrance signal ($\sigma_{\text{O},t} = 1$). We emphasize that we thus deviate from the minimum Euclidean distance rule for selecting the next landing aircraft \emph{during training}. In the validation scenarios and for the real-world deployment, the rule is imposed as described in Section \ref{sec:airspace_design}.

A training episode concludes either when the selected main agent reaches the landing zone or after 250 episode steps have passed. In the latter case, the given signaling strategy yields $200 / 250 = 80\%$ transition tuples of the main agent being connected to staying in the airspace. At the same time, $ 50 / 250 = 20\%$ of the experience relates to entering the landing zone, yielding a reasonable distribution of tuples in the replay buffer. The experience of the other $N_t -1$ in the airspace is not used for optimization since their inclusion would result in a massive underrepresentation of experience tuples associated with the VTOL zone entrance task.

\section{Results and validation}\label{sec:results}
\subsection{Training progress}\label{subsec:training_progress}
We run the LSTM-TD3 algorithm outlined in Section \ref{subsec:algorithm} for $2 \cdot 10^6$ training steps while using the hyperparameter configuration of \cite{waltz20232}. The software used for the experiments is Python 3.8.6 \citep{python3}, and the optimization of the neural networks is realized using the PyTorch deep learning library \citep{paszke2019pytorch}. Hardware-wise, the experiments run on Intel(R) Xeon(R) CPUs E5-2680 v3 (12 cores) running at 2.50 GHz. The source code of this paper is available at \cite{TUDRL}, fostering reproducibility.

\begin{figure}[htp]
    \centering
    \includegraphics[width=\textwidth]{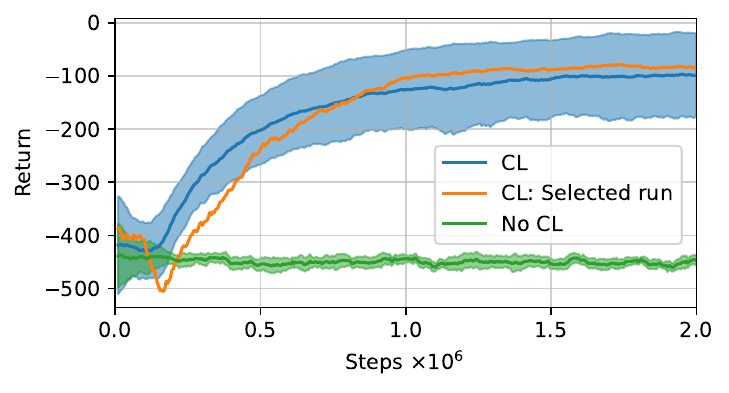}
    \caption{Training progress under different settings: 'CL' dynamically increases the number of vehicles, while 'No CL' considers 25 vehicles from the first training step. The 'CL: Selected run' is the run of the 'CL' setting which is used for the upcoming evaluations.}
    \label{fig:train_plot}
\end{figure}

Figure \ref{fig:train_plot} illustrates the progression of the test returns, which are the sum of rewards in an episode, during the training phase. The blue curve labeled 'CL' represents the curriculum learning strategy, where the number of aircraft gradually increases, as outlined in Section \ref{subsec:scenario_generation}. For comparison, we include the green curve labeled 'No CL', where 25 vehicles are generated immediately after the start of training, resulting in high initial complexity that blocks learning.

To generate the bold blue curve in Figure \ref{fig:train_plot}, we conduct ten independent experiments under the 'CL' setting. The algorithm is trained for $2 \cdot 10^6$ steps in each experiment, with five evaluation episodes conducted every 5000 training steps. The returns of these evaluation episodes are averaged and then exponentially smoothed to enhance visual clarity. Averaging the ten resulting test return curves produces the bold blue curve. Additionally, we compute a point-wise 95\% confidence interval around the bold blue line, represented by the shaded area. The same procedure is applied to the 'No CL' setting, depicted in green. Finally, the orange curve represents one of the ten experiments conducted under the 'CL' setting whose policy is analyzed in the subsequent evaluation scenarios.

The figure demonstrates that the test return of the 'CL' setting consistently improves throughout training, plateauing after approximately $1.5 \cdot 10^6$ steps. In contrast, the 'No CL' approach shows no meaningful learning progress, highlighting the importance of gradually increasing scenario complexity.

\subsection{Policy development}\label{subsec:policy_dev}
To gain deeper insights into the policy's evolution during training, we analyze its behavior at various milestones:
\begin{itemize}
    \item Policy I: After random initialization of the neural networks;
    \item Policy II: After $10^5$ steps;
    \item Policy III: After $5 \cdot 10^5$ steps;
    \item Policy IV: After training is completed.
\end{itemize}
We sequentially initialize 30 vehicles, with the details being provided in Section \ref{subsec:simstudy}. The results are presented in a publicly accessible \rd{animation}\footnote{\url{https://youtu.be/4475p8YqrEg}}. 

With Policy I, the aircraft predictably execute random maneuvers without any reasonable pattern, although there is a slight tendency to steer to the right. Remarkably, with Policy II, the algorithm discovers a suboptimal solution, manifesting as local holding patterns near the entry gates. However, this solution still results in frequent incidents. Progressing to $5 \cdot 10^5$ steps in Policy III, the algorithm demonstrates behavior closer to the final policy, albeit still susceptible to high local densities and instances of false vertiport entrances, where an aircraft enters the eVTOL zone without the corresponding signal. Lastly, in Policy IV, the aircraft successfully stay in the airspace and fly clockwise around the vertiport to achieve operational safety. Crucially, the aircraft enter the vertiport if they receive the corresponding signal, yielding a safe and efficient traffic flow. We emphasize that the clockwise behavior is learned and \emph{has not} been specified in any sense. In particular, we noticed during our experiments that whether the final policy follows a clockwise or anti-clockwise motion develops randomly, depending on the random seeds of the involved random generators. However, the traveling direction does not impact the efficiency of the solution. 

\subsection{Scenario-based validation}\label{subsec:scenario_val}
\begin{figure}[!ht]
\vspace{0.5cm}
\captionsetup[subfigure]{font=footnotesize}
    \begin{subfigure}{\linewidth}
        \centering
        \includegraphics[width=\linewidth]{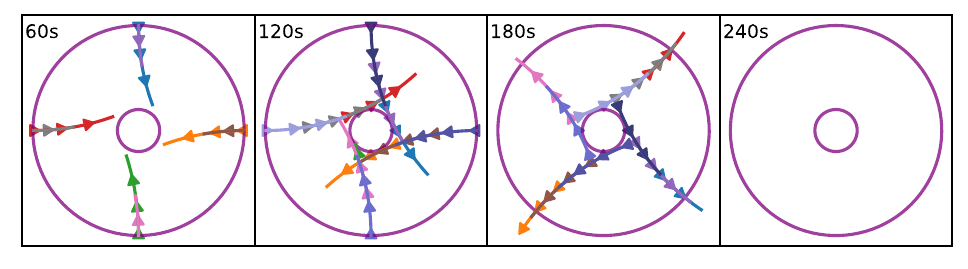}
        \caption{Policy I: After random initialization of the neural networks.}
        \vspace{0.25cm}
    \end{subfigure}
    \vspace{0.25cm}
    \begin{subfigure}{\linewidth}
        \centering
        \includegraphics[width=\linewidth]{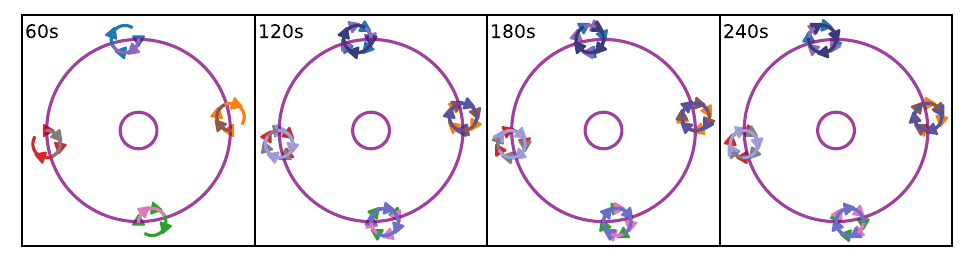}
        \caption{Policy II: After $10^5$ steps.}
    \end{subfigure}
    \begin{subfigure}{\linewidth}
        \centering
        \includegraphics[width=\linewidth]{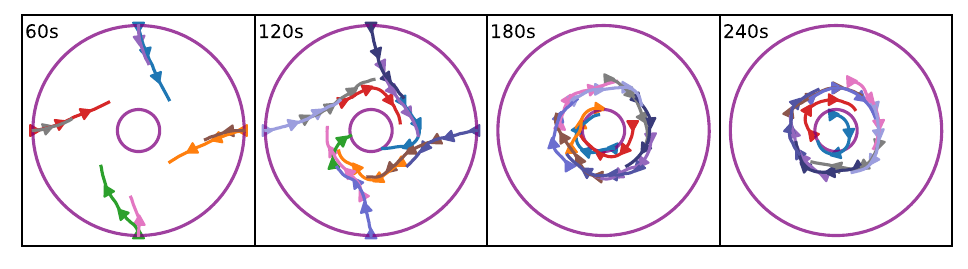}
        \caption{Policy III: After $5 \cdot 10^5$ steps.}
    \end{subfigure}
    \caption{Aircraft trajectories during the validation scenario with Policies I, II, and III. The triangles of a trajectory are $\unit[20]{s}$ apart.~Considering the speed of $\unit[13]{m/s}$, the $\unit[20]{s}$ correspond to approximately $\unit[260]{m}$.}
    \label{fig:policies_val}
\end{figure}

We further illustrate the learned behavior in a scenario consisting of three inbound waves of aircraft. In each wave, four aircraft enter the airspace simultaneously, one through each entrance gate, with the heading pointing toward the eVTOL zone midpoint. The vehicle waves are $\unit[30]{s}$ apart and we assume each vehicle has a constant speed of $\unit[13]{m/s}$. For comparison purposes, we include the resulting trajectories of Policies I, II, and III in Figure \ref{fig:policies_val}, reflecting their limitations outlined in Section \ref{subsec:policy_dev}. 

\begin{figure}[!ht]
    \centering
    \includegraphics[width=\textwidth]{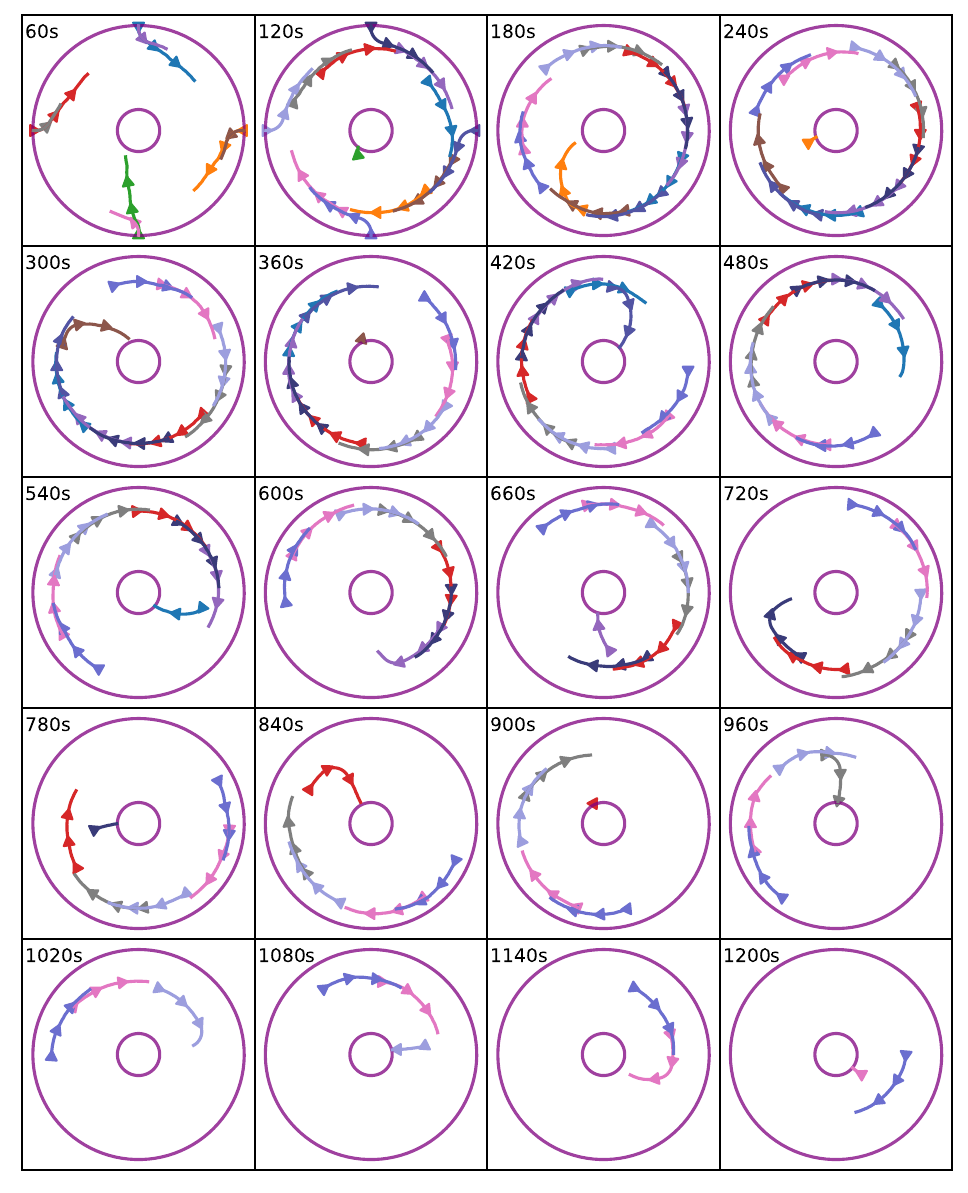}
    \caption{Aircraft trajectories during the validation scenario with Policy IV. The triangles of a trajectory are $\unit[20]{s}$ apart.~Considering the speed of $\unit[13]{m/s}$, the $\unit[20]{s}$ correspond to approximately $\unit[260]{m}$.}
    \label{fig:val_traj}
\end{figure}

\begin{figure}[!ht]
    \centering
    \includegraphics[width=\textwidth]{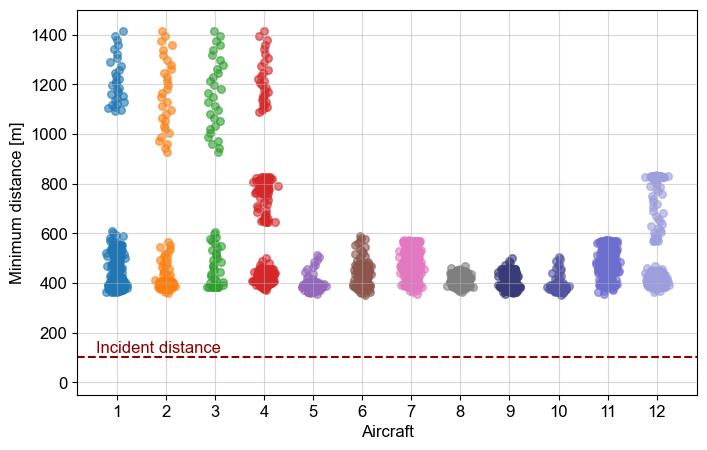}
    \caption{Empirical distribution of the minimum distance to other vehicles of each aircraft with Policy IV.}
    \label{fig:val_dist}
\end{figure}

Figure \ref{fig:val_traj} shows the trajectories of Policy IV, illustrating the clockwise motion around the vertiport. According to Figure \ref{fig:val_dist}, which displays the distribution of the minimum distances to other vehicles, there are no accidents or incidents during the scenario. The lowest aircraft distances are above $\unit[300]{m}$, indicating high operational safety. The clusters in the point clouds of vehicles 1, 2, 3, 4, and 12 can be explained by aircraft entering and leaving the airspace, possibly creating a jump in the minimum distance to other vehicles.

\subsection{Simulation study}\label{subsec:simstudy}
We perform a thorough simulation study to quantitatively evaluate the Policy IV's performance.~In particular, we spawn in total ${N \in \{5, 10, 15, 25, 30\}}$ vehicles per episode, and the \rd{animation} provided in Section \rd{\ref{subsec:policy_dev}} illustrates the case for $N = 30$. The aircraft appear at a random gate with a $\unit[15]{s}$ time gap. We disturb the heading of each aircraft, which points toward the eVTOL zone midpoint, by a realization of $\mathcal{U}(-20,20) \cdot 1^\circ$. In addition, we randomly sample the aircraft speeds from $\mathcal{U}(10,16) \cdot \unit[1]{m/s}$. \rd{We repeat the experiment 30 times for each $N$, tracking the number of accidents and incidents as safety metrics. Additionally, we monitor the number of false vertiport entrances. To quantify operational efficiency, we measure the average airspace time, defined as the duration from entering the airspace to entering the vertiport. This metric is the sum of two components: The average time to signal, which is the time it takes for an aircraft to receive the entrance signal, and the average entrance time, which is the time required to enter the vertiport after receiving the signal.}

\begin{figure}[!ht]
    \centering
    \includegraphics[width=\textwidth]{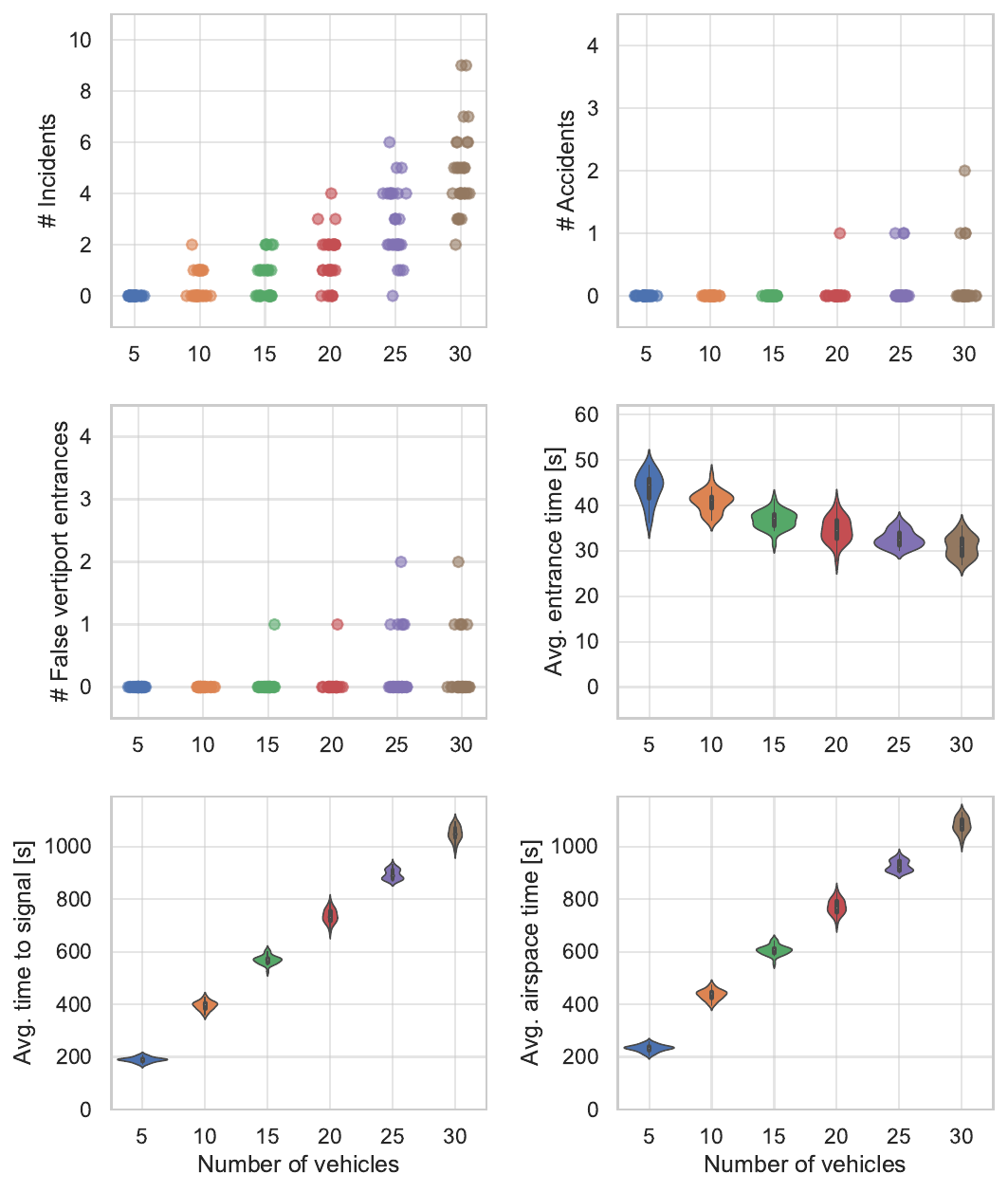}
    \caption{\rd{Results of the simulation study with Policy IV, when no safety check for airspace entrance is performed.}}
    \label{fig:simstudy_results}
\end{figure}

Analyzing Figure \ref{fig:simstudy_results}, the aircraft consistently achieve moderate operational safety, with accidents occurring in only 8 out of the 180 conducted runs. These cases are exclusive to scenarios characterized by high traffic densities involving 20 or more vehicles. While slightly elevated, the incident rates remain in the single digits for the analyzed scenarios. 

A more profound understanding of these events is gleaned from the \rd{animation} presented in Section \ref{subsec:policy_dev}. The root cause of most incidents lies in aircraft entering the space, particularly when other vehicles are near the gate. In an operational environment, this problem is solved by pre-arrival management, which informs aircraft already in the cruise phase about arrival times and provides advisories for in-time arrivals. Recognizing this observation, we conducted a secondary simulation study, incorporating a simple additional safety check for airspace entrance. Specifically, we examined whether any aircraft within the airspace maintained a Euclidean distance of less than $\unit[300]{m}$ to the respective gate. If affirmative, an alternative gate was selected for entrance; if all gates were occupied, no entrance is possible.

\begin{figure}[!ht]
    \centering
    \includegraphics[width=\textwidth]{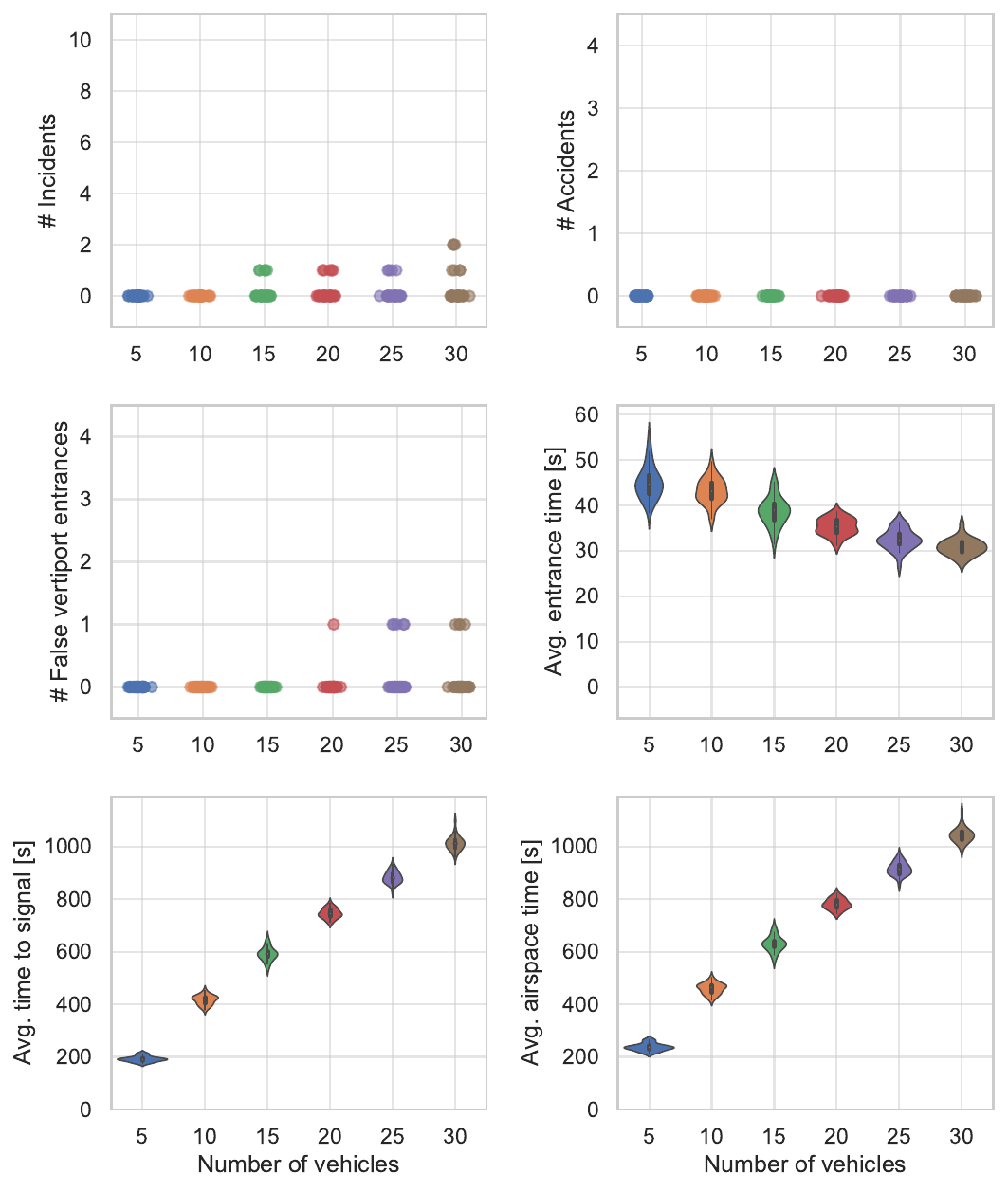}
    \caption{\rd{Results of the simulation study with Policy IV, when a safety check for the airspace entrance is implemented.}}
    \label{fig:simstudy_results_check}
\end{figure}

The outcomes of this refined simulation are illustrated in Figure \ref{fig:simstudy_results_check}, revealing a substantial reduction in incidents and, notably, a complete absence of accidents. This underscores the effectiveness of the introduced safety check, which is straightforward to implement in real-world scenarios. On this basis, the Policy IV ensures high operational safety, providing a robust method for safe and efficient terminal airspace operations.

\rd{In addition, Figures \ref{fig:simstudy_results} and \ref{fig:simstudy_results_check} show almost no false vertiport entrances except in cases with higher traffic density. Moreover, the airspace time increases linearly with the number of vehicles from approximately \unit[220]{s} for $N = 5$ to over \unit[1000]{s} for $N = 30$, which is primarily driven by the linear increase in the time to signal. This observation is expected since higher traffic density causes vehicles to wait longer for the entrance signal due to the limited capacity of the vertiport. Notably, the average entrance time decreases from approximately \unit[45]{s} for $N=5$ vehicles to slightly above \unit[30]{s} for $N=30$ vehicles. This reduction is due to the higher traffic density, which forces vehicles to spread out more broadly in the available airspace. Consequently, the distance of the closest vehicle to the eVTOL zone is reduced, leading to a quicker entrance time. However, this decrease in entrance time is negligible compared to the increase in the time to signal.}

\begin{figure}[htp]
    \centering
    \includegraphics[width=\textwidth]{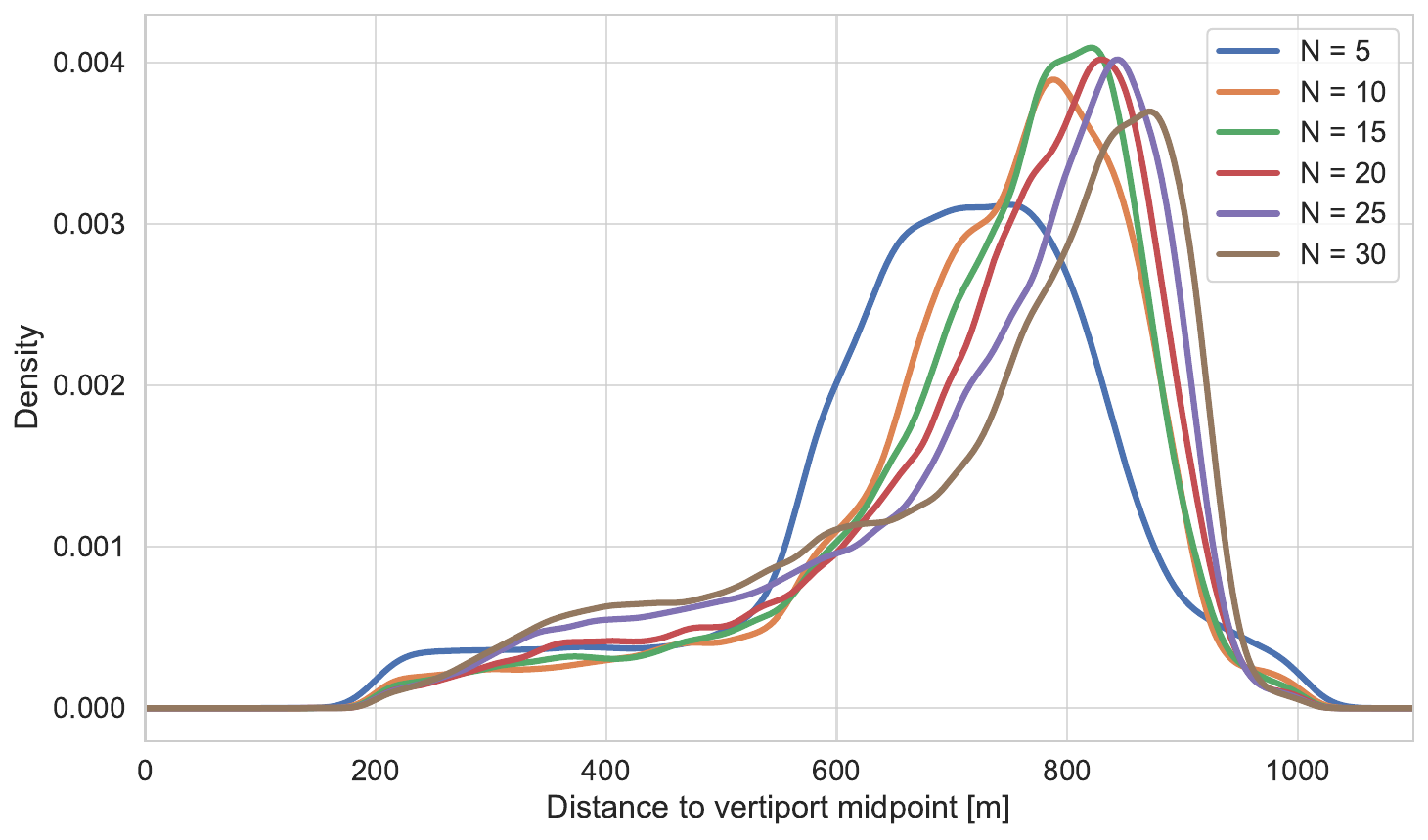}
    \caption{Spatial distribution of the aircraft during the simulation study with Policy IV, when no entrance check has been performed.}
    \label{fig:simstudy_spatial}
\end{figure}

Lastly, Figure \ref{fig:simstudy_spatial} shows the spatial distribution of the aircraft depending on the number of vehicles. The distribution is estimated using a kernel density estimator with a Gaussian kernel, leveraging the rule of thumb of \cite{silverman2018density} for bandwidth selection. The mode of the distribution is further away from the vertiport with an increasing number of vehicles, reflecting the need to spread out through the airspace. In addition, with decreasing $N$, the vehicles traveling toward the VTOL zone have an increasing impact on the distribution, constituting another factor that shifts the mode toward the landing zone.

\section{\rd{Robustness analysis}}\label{sec:robustness}
\rd{In this section, we explore several methods to assess the robustness of Policy IV. First, we introduce Gaussian noise to the aircraft positions during decision-making to account for the limitations of practical sensing devices. Second, we modify the inbound air traffic to follow a Poisson distribution, reflecting common practices in conventional air traffic modeling. Third, we test Policy IV on small-scale drones to evaluate its transferability to real-world applications. Lastly, we include a discussion of the findings.}

\subsection{\rd{Positional noise}}\label{subsec:noise}
\rd{The location estimation of eVTOL vehicles is likely to rely on satellite-based radio navigation systems such as the Global Positioning System. After a vehicle determines its position, it can broadcast this information to surrounding aircraft, which use it as a basis for their distributed decision-making. However, practical sensing devices can introduce inaccuracies due to signal obstructions from surrounding objects, variations in atmospheric conditions, or limited receiver quality \citep{thin2016gps}, possibly negatively impacting the overall operation.}

\rd{We investigate how such positional noise affects the efficiency and safety of Policy IV. For vehicle \( i \), we denote the ground-truth position vector at time \( t \) as \( p_{i,t}^{\text{G}} = \left(n_{i,t}^{\text{G}}, e_{i,t}^{\text{G}}\right)^\top \), where \( n_{i,t}^{\text{G}} \) and \( e_{i,t}^{\text{G}} \) are the north and east components, respectively. Based on this, we construct the noisy position vector \mbox{\( p_{i,t}^{\text{N}} = \left(n_{i,t}^{\text{N}}, e_{i,t}^{\text{N}}\right)^\top \)} with north component \( n_{i,t}^{\text{N}} \) and east component \( e_{i,t}^{\text{N}} \) as follows:
\begin{equation}
      p_{i,t}^{\text{N}} = p_{i,t}^{\text{G}} + \epsilon_{i,t},  
\end{equation}
where $\epsilon_{i,t} \sim \mathcal{N}\left( \begin{pmatrix} 0 \\ 0 \end{pmatrix}, \begin{pmatrix} \sigma_{\text{N}}^2 & 0 \\ 0 & \sigma_{\text{N}}^2 \end{pmatrix} \right)$ is a realization of a two-dimensional zero-mean Gaussian random vector, with \(\sigma_{\text{N}}\) being the standard deviation. The noise is assumed to be uncorrelated over time. In the following, the computations of vehicle observations (see Section \ref{subsec:obs_act_space}), including distances, angles, and risk measures, rely entirely on the noisy positions $p_{i,t}^{\text{N}}$.}

\rd{In particular, we consider three distinct noise intensities: Small noise with $\sigma_{\text{N}} = \unit[10]{m}$, medium noise with $\sigma_{\text{N}} = \unit[20]{m}$, and large noise with $\sigma_{\text{N}} = \unit[100]{m}$. We replicate the simulation study of Section \ref{subsec:simstudy} for each noise level to conduct a quantitative evaluation. The results are displayed in \ref{app:pos_noise}. In addition, we provide \rd{an animation}\footnote{\rd{\url{https://youtu.be/nVCsP6YOm58}}} to visualize the effect of the positional noise. In the \rd{animation}, the ground truth positions are represented in darker colors, while the noisy positions used for action computation oscillate in lighter colors around the ground truth positions.}

\rd{The \rd{animation} and simulation results clearly demonstrate Policy IV's robustness against positional measurement errors. The flight behavior remains smooth, maintaining high safety and efficiency levels for small to medium noise levels. We primarily attribute this strong performance to the architecture of the neural network in Figure \ref{fig:LSTMRecTD3}. The temporal recurrent component effectively manages partial observability, ensuring Policy IV's noise resilience. The incident frequency only increases in the large noise case, where accidents are also observed. However, we emphasize that $\sigma_{\rm N} = \unit[100]{m}$ constitutes an extreme scenario with limited practical relevance to truly test the limits of Policy IV. To summarize, our DRL-based approach offers strong robustness against measurement errors from real-world sensing devices.}

\subsection{\rd{Distribution of inbound traffic}}
\rd{The validation scenario of Section \ref{subsec:scenario_val} considered several waves of aircraft, in which each wave consisted of four entering aircraft, one per gate. Furthermore, the simulation studies of Sections \ref{subsec:simstudy} and \ref{subsec:noise} assumed single aircraft entering at a random gate with a $\unit[15]{s}$ time gap to achieve a steady increase in traffic density. While these approaches are suitable for investigating the performance of our DRL policy, we acknowledge that inbound traffic in conventional air traffic is frequently modeled via a Poisson distribution \citep{lancia2020predictive}. Therefore, we investigate Policy IV's robustness against the distribution of inbound traffic via a scenario with Poisson-distributed arrivals. In particular, we assume all aircraft to enter the airspace via the south gate, representing a clustered arrival through one flight corridor. We assume four clusters with a $\unit[120]{s}$ time gap between consecutive clusters. Each cluster's number of aircraft is sampled from a Poisson distribution with parameter $\lambda = 5$ so that, in expectation, 20 aircraft are generated. The aircraft inside one cluster enter the airspace with a small time gap of $\unit[10]{s}$. Two independent runs of the scenario are shown in \rd{an animation}\footnote{\rd{\url{https://youtu.be/FxuDZTx2kWw}}}.}

\rd{As the \rd{animation} illustrates, the distribution of inbound traffic has minimal impact on operational performance, provided that the vehicles' entrance does not immediately cause an incident, which can be mitigated with a simple safety check as discussed in Section \ref{subsec:simstudy}. In the \rd{animation}, the arriving Poisson sequence of vehicles quickly disperses in the airspace as they begin to spread out. Consequently, Policy IV demonstrates strong robustness to variations in the distribution of inbound traffic.}

\subsection{Sim-2-Real transfer}
\begin{figure}
    \centering
    \includegraphics[width=0.4\textwidth]{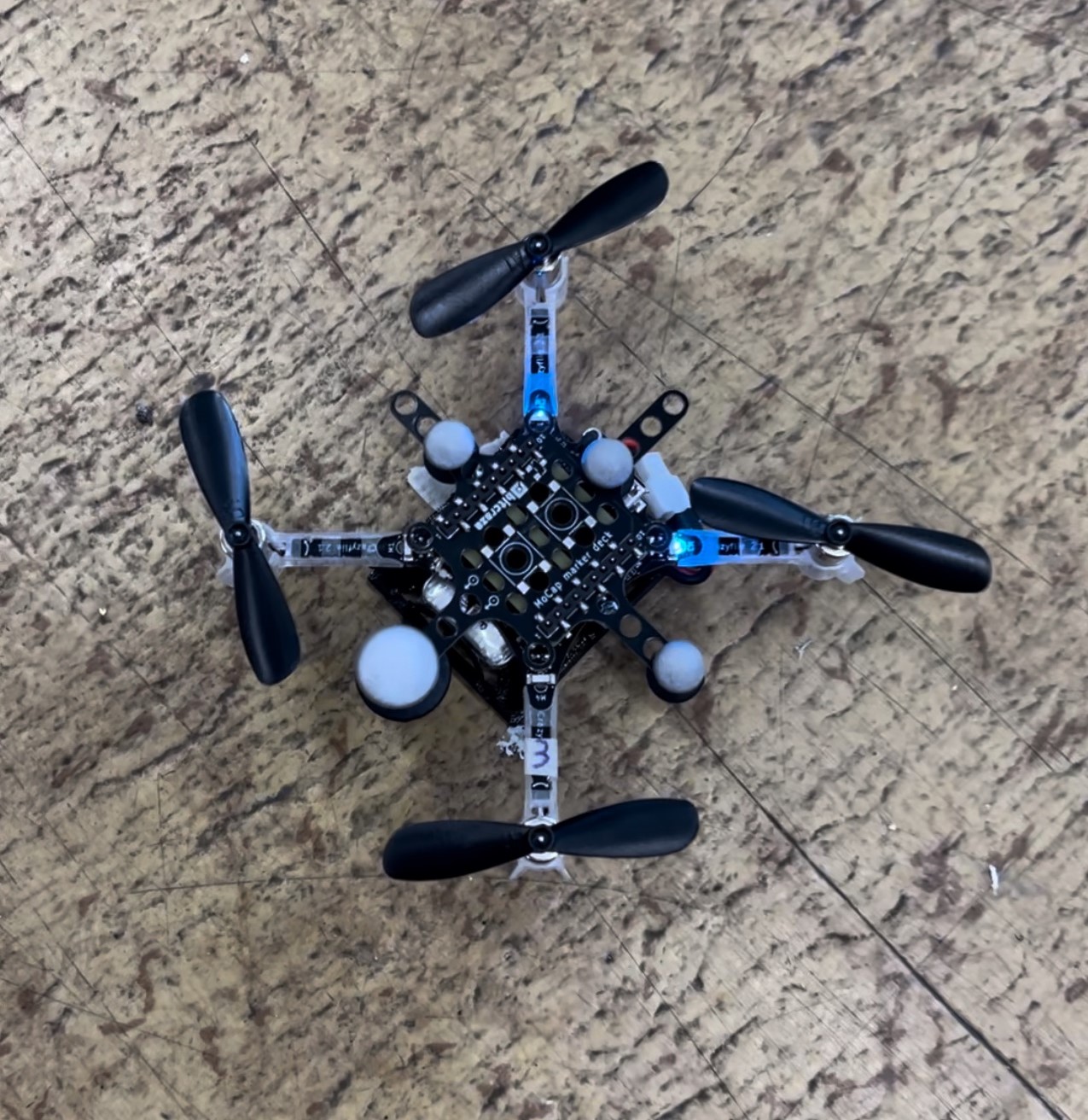}
    \caption{One of the Crazyflie drones used for the real-world experiments.}
    \label{fig:CF}
\end{figure}

We demonstrate the real-world applicability of our DRL-based control approach by deploying Policy IV on five Crazyflie drones \citep{giernacki2017crazyflie}. The results are showcased in a video\footnote{\url{https://youtu.be/wbHUxARpHzM}}. As depicted in Figure \ref{fig:CF}, the Crazyflie is a nano quadcopter used as an experimental platform for research and education. \rd{One drone weighs approximately \unit[27]{g} and offers battery for up to \unit[7]{min} of flight time before recharging. The technical setup is as follows: The drones' positions are estimated using a camera-based motion capture system, while the DRL-related computations are performed on a ThinkPad-P15v-Gen-3 notebook with an Intel i7-12700H processor. The ROS2-based Crazyswarm2 \citep{crazyswarm2} testbed acts as middleware for communicating with the drones. In particular, the notebook receives information from the motion capture system and computes the high-level control commands for each drone using the DRL policy. The latter provides heading commands, from which we compute a desired position for each drone assuming a constant velocity; see Section \ref{sec:airspace_design}. These desired positions are sent to the drones, which are equipped with an STM32F405 onboard microcontroller to derive low-level actuator commands. More precisely, we use a differential flatness-based low-level controller \citep{mellinger2011minimum} to govern the drone actuators.}

In our experimental setup, five drones are initially equidistantly spaced in the airspace. \rd{After a drone enters the vertiport, it is out of the scope of our DRL policy, and we hard-code the respective landing position to ensure there are no collisions.} The radius of the airspace is set to $\unit[1.2]{m}$, and the drones maintain a velocity of approximately $\unit[0.15]{m/s}$. The video illustrates a stable and efficient operation, with the drones maintaining a safe distance and entering the vertiport sequentially. Crucially, we did \emph{not} retrain the policy to accommodate the reduced size of the environment compared to the training scenarios. Since the agents' observation space is properly normalized (see Section \ref{subsec:obs_act_space}), adjusting the normalizing constants to match the smaller scale of the experiment suffices. The successful zero-shot adaptation of our policy is remarkable, considering that Sim-2-Real transfer often poses a significant challenge for RL algorithms \citep{zhao2020sim, muratore2019assessing}.

\subsection{\rd{Discussion}}
\rd{The prior experiments and robustness analyses underscore the generalizability and practical usability of our DRL-based approach. In particular, we showcase that Policy IV is robust to partial observability in the form of noise and real-world disturbances. From a policy maker's perspective, these observations make a strong case for DRL in contrast to traditional heuristic policies or local optimization routines. DRL excels in adapting to complex and dynamic environments where conventional methods may struggle due to their static nature. It can learn to handle uncertainty and stochasticity inherent in real-world environments more effectively \citep{kober2013reinforcement, mnih2015human}. Furthermore, DRL can consider multiple objectives simultaneously, in our case, minimizing collisions, guaranteeing vertiport entrance, staying in the airspace, and selecting comfortable actions. These trade-offs are often difficult to explicitly encode in a heuristic policy without sacrificing simplicity or performance. Lastly, from a computational point-of-view, our DRL approach has an extremely low inference time, allowing for real-time distributed decision-making with an arbitrary number of vehicles. To summarize, while heuristic policies have their place, particularly in well-defined and stable environments, DRL provides a promising approach for addressing the challenges and uncertainties inherent in future UAM systems.}

\section{Conclusion}\label{sec:conclusion}
Urban air mobility holds the promise of transforming the transportation landscape, alleviating traffic congestion, and promoting sustainability. The terminal arrival phase in UAM operations is of paramount importance from a safety standpoint. This paper introduces a self-organized arrival system using deep reinforcement learning. The proposed method allows for decentralized action selection based on local observations, mitigating the risk associated with a single point of failure in practical operations. Our approach demonstrates safe and efficient traffic flow, with real-world experiments on small-scale drones validating its practical applicability.

Several avenues for future research are identified to build upon this work. Methodologically, we operated under the assumption of a non-cooperative setting where communication between vehicles is not considered. Relaxing this assumption could leverage communication to share intentions among vehicles, potentially further enhancing the safety of operations. Furthermore, in the current framework, a simple rule dictates the order of aircraft entering the landing zone. The integration of communication could render this rule obsolete, empowering agents to decide their entry sequence autonomously. Finally, a next step is the deployment of this methodology on full-scale eVTOL vehicles, providing further validation in practical operational scenarios.

\section{Acknowledgements}
The authors thank Peng Huang and the team of the Barkhausen Institute for the great assistance with the Crazyflie drones. Moreover, the authors acknowledge the Center for Information Services and High Performance Computing at TU Dresden for providing the resources for high-throughput calculations. This research did not receive any specific
grant from funding agencies in the public, commercial, or
not-for-profit sectors.

\bibliographystyle{elsarticle-harv} 
\bibliography{bib}

\newpage
\appendix

\setcounter{table}{0}
\setcounter{figure}{0}
\section{\rd{Results for noise robustness}}\label{app:pos_noise}
\begin{figure}[!ht]
    \centering
    \includegraphics[width=0.925\textwidth]{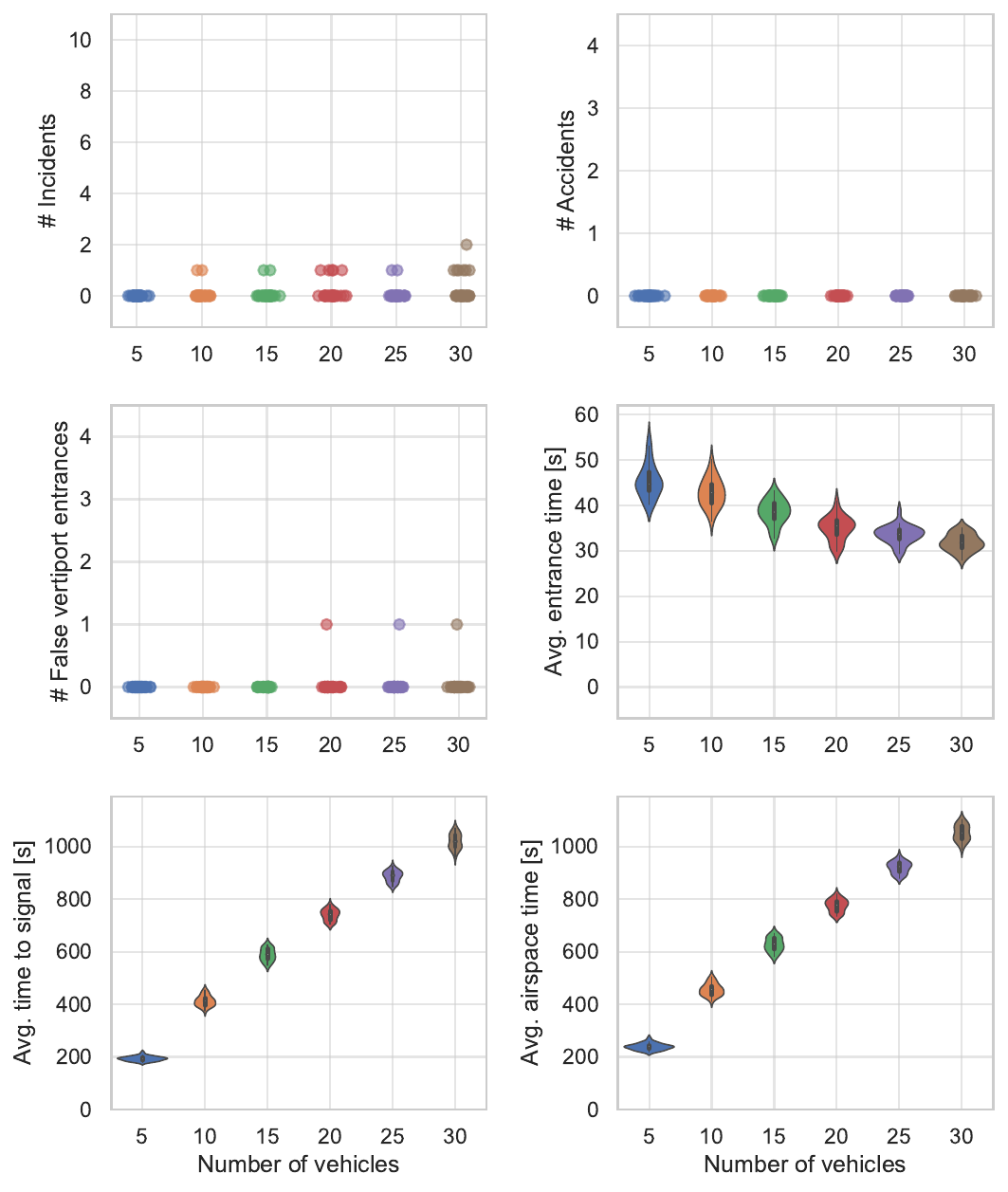}
    \caption{\rd{Results of the simulation study with Policy IV under observations with small positional noise ($\sigma_{\rm N} = \unit[10]{m}$).}}
    \label{fig:simstudy_noise_10}
\end{figure}

\begin{figure}[!ht]
    \centering
    \includegraphics[width=\textwidth]{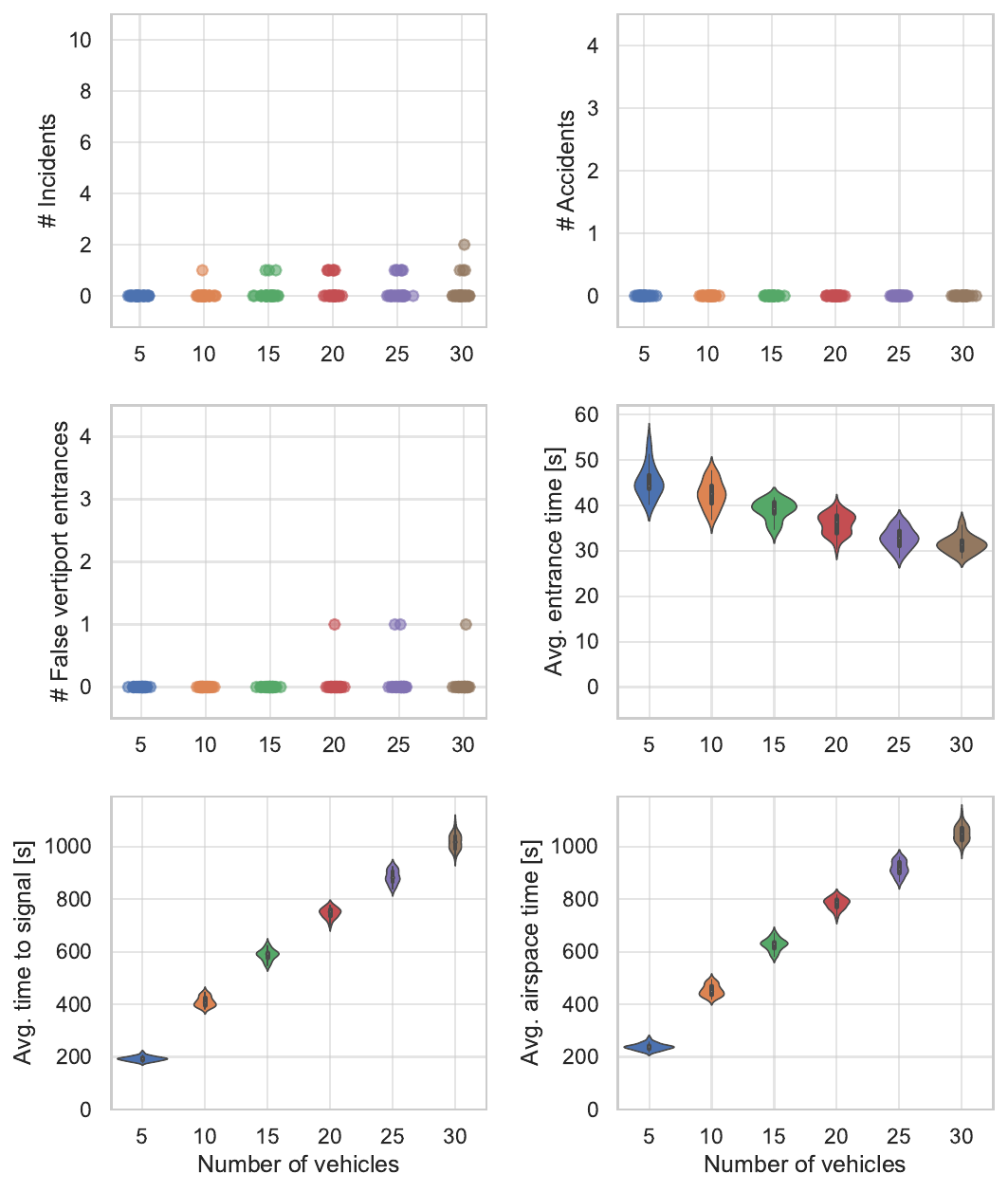}
    \caption{\rd{Results of the simulation study with Policy IV under observations with medium positional noise ($\sigma_{\rm N} = \unit[20]{m}$).}}
    \label{fig:simstudy_noise_20}
\end{figure}

\begin{figure}[!ht]
    \centering
    \includegraphics[width=\textwidth]{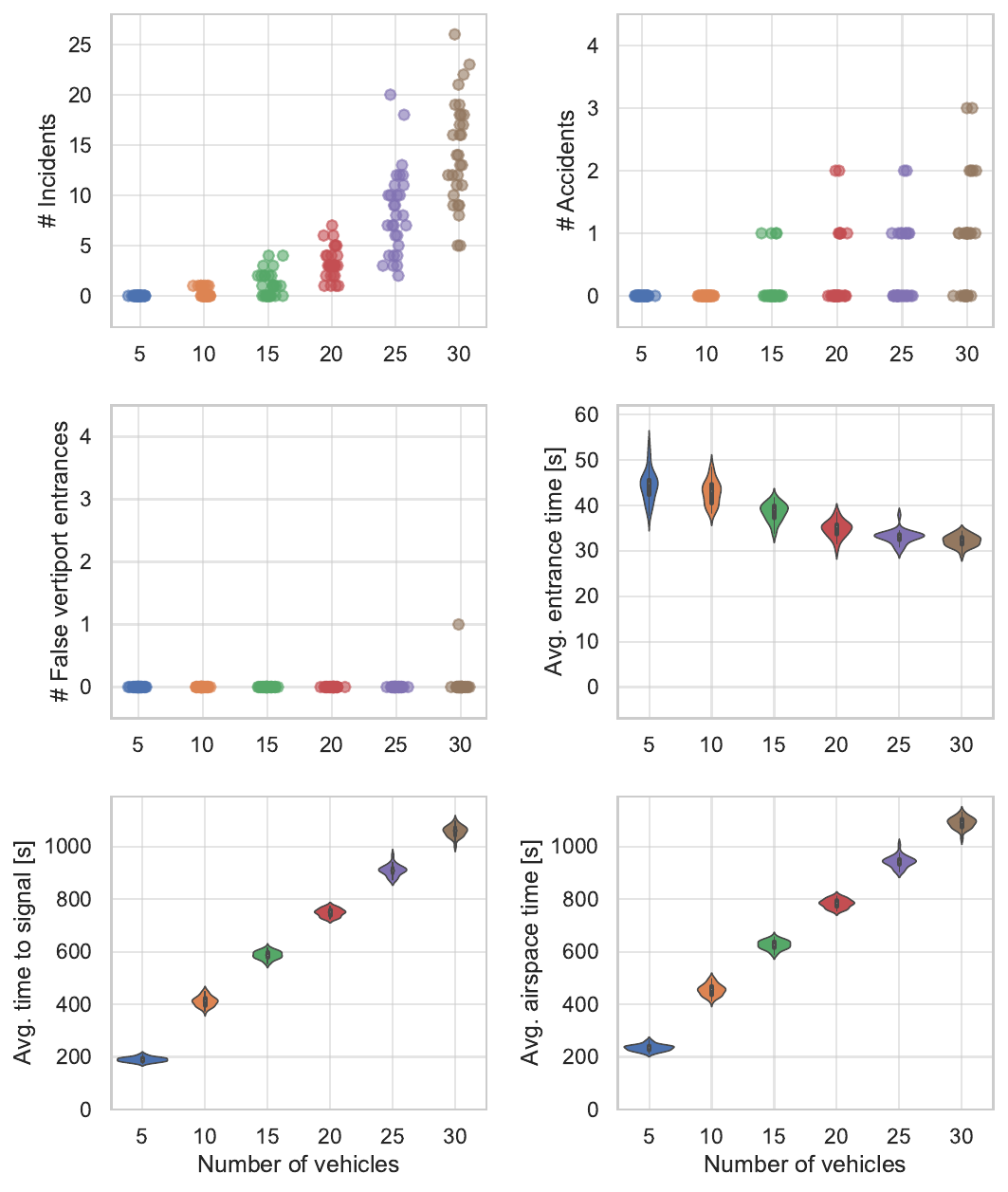}
    \caption{\rd{Results of the simulation study with Policy IV under observations with large positional noise ($\sigma_{\rm N} = \unit[100]{m}$).}}
    \label{fig:simstudy_noise_100}
\end{figure}

\rd{Figures \ref{fig:simstudy_noise_10} - \ref{fig:simstudy_noise_100} show the results of the simulation study under noisy observations from Section \ref{subsec:noise}. The safety check for airspace entrance is implemented as outlined in Section \ref{subsec:simstudy}.}

\end{document}